\documentclass[runningheads]{llncs}
\usepackage{graphicx}
\usepackage{bbm}
\usepackage{amsfonts}       
\usepackage{booktabs}       
\usepackage{multicol}
\usepackage[table,xcdraw]{xcolor}
\usepackage{amsmath,amssymb}
\usepackage[colorinlistoftodos]{todonotes} 

\usetikzlibrary{positioning}
\tikzset{main node/.style={circle,fill=blue!20,draw,minimum size=1cm,inner sep=0pt},}

\usepackage[caption=false]{subfig}
\usepackage{wrapfig}

\newcommand{\CC}{\mathcal{C}}

\newcommand{\OO}{\mathcal{O}}
\newcommand{\RR}{\mathbb{R}}

\newcommand{\mat}[1]{ {\ensuremath{\mathsf{#1} }}}

\def\matX{\mat{X}}

\newcommand{\y}{\mathbf{y}}

\begin{document}
%
\title{Team voyTECH: User Activity Modeling with Boosting Trees}
%
%

\author{
  Immanuel Bayer\inst{1}\and Anastasios Zouzias\inst{2}
}

\authorrunning{I. Bayer and A. Zouzias}
%
\institute{Palaimon GmbH \\ Berlin, Germany \\
\email{immanuel.bayer@palaimon.io}\\
\url{https://palaimon.io}
\and
Huawei Technologies\\
Zurich Research Center \\
Switzerland \\
\email{anastasios.zouzias@huawei.com}
}
\maketitle              
%

\begin{abstract}

This paper describes our winning solution for the ECML-PKDD ChAT Discovery Challenge 2020.
We show that whether or not a Twitch user has subscribed to a channel can be well predicted by modeling
user activity with boosting trees. We introduce the connection between target-encodings and boosting trees in the context of high cardinality categoricals and find that modeling user activity is more powerful then direct modeling of content when encoded properly and combined with a gradient boosting optimization approach.

\end{abstract}
%
\section{The Competition}
%
The task of the ECML-PKDD ChAT Discovery Challenge 2020~\cite{kobs2020towards} is to predict whether or not a Twitch user has subscribed to a channel (binary classification task) given the list of messages he has posted on this and other channels.

The dataset consists of 700 million public Twitch comments taken from english channels that are published during the month of January 2020 along with metadata. The training data consists of over 29 million and the test dataset of 90,000 channel-user combinations and their comments. In more detail, each input training sample consists of the channel-id, the user-id, and a list of time-stamped comments from this user, including the specific game played in this channel at this particular time.
%
\subsection{Competition Challenges}
%
%
The ChAT competition presents two peculiarities. The first challenge is that only half of the users in the test set have no history which requires special attention when extracting users and channels features. This challenge draws similarities with the cold start problem in recommendation systems~\cite{recommender:cold_start}. The second challenge is related to the sampling distribution of the test set (leader-board). More precisely, the entire spectrum of user/channel activity levels (low, normal, high) is weighted equally across all groups which is vastly different than their frequencies in the training set (see Table~\ref{tab:split}). Namely, for each out of $9$ combinations of user activity levels (low, normal, high) and channel activity levels (low, normal, high), 10,000 channel-user pairs are sampled uniformly (i.e., one channel-user activity group is where the user is of low activity and the channel is of normal activity). Hence, in total 90,000 test samples are generated. In Table~\ref{tab:dataset:stats}, we outline statistics of the dataset within the activity groups. 
\begin{table}[ht]
\caption{Statistics of the training dataset per channel-user activity group. `u\_low-c\_normal` corresponds to low user and normal channel activity group.}\label{tab:dataset:stats}
\label{tab:split}
\scriptsize
\centering
\begin{tabular}{|c|c|c|c|c|c|c|}
\toprule
\textbf{group}               & \textbf{users} & \textbf{channels} & \textbf{pairs} & \textbf{subscribed} & \textbf{\% pairs} & \textbf{\% subscribed} \\
\midrule
\textbf{u\_low-c\_normal}    & 141K          & 40K              & 144K           & 5,696                   & 0.0049          & 0.0397               \\ \hline
\textbf{u\_low-c\_low}       & 10K           & 7,7K               & 10K           & 611                    & 0.0003          & 0.0595               \\ \hline
\textbf{u\_normal-c\_normal} & 480K          & 67K              & 562K           & 35,021                  & 0.0190          & 0.0622               \\ \hline
\textbf{u\_low-c\_high}      & 2,153K        & 34K              & 2,359K        & 181K                 & 0.0798          & 0.0768               \\ \hline
\textbf{u\_normal-c\_low}    & 46K           & 23K              & 47K            & 3651                   & 0.0016          & 0.0770               \\ \hline
\textbf{u\_normal-c\_high}   & 3,508K        & 36K              & 8,740K        & 683K                & 0.2958           & 0.0782               \\ \hline
\textbf{u\_high-c\_high}     & 1,911K        & 36K              & 16,045K        & 1,314K               & 0.5432          & 0.0819               \\ \hline
\textbf{u\_high-c\_low}      & 77K           & 31K              & 99K            & 8,498                   & 0.0033          & 0.0858               \\ \hline
\textbf{u\_high-c\_normal}   & 663K          & 73K              & 1,531K        & 135K                 & 0.0518          & 0.0886               \\
\bottomrule
\end{tabular}
\end{table}

%
\subsection{Contributions}
%
The main contributions of this paper are:
\begin{itemize}
\item The detailed presentation of the winning solution of the Discovery Challenge 2020 including training and evaluation setup.
\item Introduction of the connection between target encodings and boosting trees in the context of high cardinality categoricals.
\item Additional experiments on the competition dataset that examine the critical modeling decisions of our solution.
\end{itemize}
%
%
\subsection{User Activity Modeling}
%
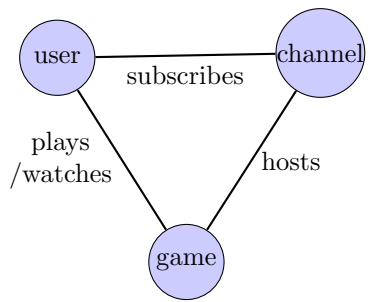
\begin{wrapfigure}{h}{0.4\textwidth}
  \vspace{-40pt}
  \begin{center}
\begin{tikzpicture}
    \node[main node] (1) {game};
    \node[main node] (2) [above left = 2.cm and 1.cm of 1]  {user};
    \node[main node] (3) [above right = 2.cm and 1.cm of 1] {channel};
    \path[draw,thick]
    (1) edge node [left, align=center]{plays\\/watches} (2)
    (2) edge node [below]{subscribes} (3)
    (3) edge node [right]{hosts} (1);
\end{tikzpicture}
  \end{center}
  \vspace{-20pt}
  \caption{Interactions}
  \vspace{-20pt}
\label{fig:interactions}
\end{wrapfigure}
Our approach is based on the assumption that modeling user activity is more important than specific content (e.g. message text).
User activity is modeled as interactions between the user and key objects of the system she interacts with (e.g. channels and games).
This approach naturally leads to a high dimensional categorical feature/variable representation that has been well studied in the context of recommender systems, click-through-rate predictions and similar industrial applications. It is also closely related to the concept of graph-based relational features~\cite{bayer2015graph}.
Our experimental results (Section \ref{sec:winning-model}) indicate that the interactions illustrated in Figure \ref{fig:interactions} together with features describing the quantity of user activity (e.g. days active, number of frequently used channels) have strong predictive power.
Introducing game-id as a high level object is especially important for the 50\% test set user without history (cold-start).
For a cold-start user, their most frequent game-id can effectively proxy their user-id (more details in Section~\ref{sec:feature-engineering}).
Before presenting details of our solution (Section~\ref{sec:experiments}), we first introduce the concept of target encoding to motivate our choice to combine high dimensional feature representations with boosting tree models.

\section{High Cardinality Categoricals and Boosting Trees}
%
In this section, we discuss in detail the interaction between (high cardinality) categoricals, mean target encodings and boosting trees which is at the core of our winning solution in form of the popular CatBoost library that we used to implement our models~\cite{catboost}.
Several user and channel categorical features are present in the dataset such as which game has been played and activity levels for user, games, and channels. By computing the interaction features between user and channel representations, several categoricals with high cardinality are extracted. Due to their sparsity, such high cardinality categoricals pose several challenges in modeling and, in general, can lead to poor generalization performance. A popular class of models to handle such a semi-structured datasets containing high cardinality categoricals are Gradient Boosting Trees~\cite{book:elem_stat_learning} and in particular the CatBoost library. The winning solution is based on a single CatBoost model. Model ensembles are likely to further improve our results, but were skipped due to time restrictions.
%
\subsection{Categorical Encodings in Models}\label{sec:models}
%
The handling of categorical features usually happens during the feature engineering phase, where the modeler has the freedom to arbitrarily transform or extract the input features before those are fed into a model. However, models exist that can handle categorical features automatically, i.e., the modeler simply specifies the features that should be handled as categoricals without any further pre-processing required. The user of such models is only able to adjust the predefined categorical encoding process with input through hyper-parameters. For example, hyper-parameters for categorical features include `perform one-hot-encoding if cardinality of any categorical is less than a threshold`, `perform hash encoding with specified number of hashing dimensions` to name a few. 
Here, we summarize a few recently proposed models that handle categorical features as part of the model definition. Two gradient boosting tree implementations, Microsoft's LightGBM~\cite{lightgbm} and Yandex's CatBoost~\cite{catboost}, allow the user to specify which features should be handled as categoricals by the models. The \emph{h2o.ai} implementation of Random Forests handles categoricals out of the box. Neural networks can provide an embedding layer to handle categoricals as an additional layer of a neural network, see Keras embedding layer or the so-called `entity embeddings`~\cite{encoding:nn:entity}. LightGBM splits a categorical feature by partitioning its categories into 2 subsets. If the categorical feature has $k$ levels, there are $2^{(k-1)} - 1$ possible partitions. However, there is an efficient $\OO(k\log (k))$ time solution for regression trees~\cite{lightgbm:cat:split}. The basic idea is to sort the categories according to the training objective at each split. 
CatBoost is a gradient boosting tree implementation that applies a regularized mean target encoding on the top-level tree split as a preprocessing step. Such preprocessing could be considered sub-optimal, at least for the case of trees with large depth~\cite{book:breiman1984classification}. Although the CatBoost approach might result in sub-optimal greedy binary splits, CatBoost requires less operations per tree split and offers a very efficient and optimized implementation. The efficiency if based on the property that regularized mean target encoding values are computed only once compared to the optimal greedy approach where the mean target encodings have to be maintained or computed on every tree split.
In the following section, we provide more background on the fundamentals of CatBoost and, in particular, its connection to mean target encodings since mean target encodings are the core design principle behind CatBoost.
%
\subsection{Target Encodings}
%
In this section, we setup the framework of feature extraction from categoricals that is usually called \emph{target encodings} from machine learning practitioners.
We denote $m$ samples with $n$ features by a $m\times n$ design matrix $\matX$ with column coefficients that are either numericals (in $\RR$) or categoricals\footnote{Throughout this paper, a feature is an input variable/predictor that is used for prediction. A categorical feature (or categorical) is a feature with a domain that is a fixed set without an explicit ordering. The elements of a categorical are referred as \emph{levels}. For example, postal code, favorite color, city or country of a specific individual are examples of categoricals.}. In other words, the $j$-th column of $\matX$ is in $\RR^m$ or $\CC_{j}^m$ for a set of elements of categoricals $\CC_{j}$. Moreover, we denote by $\matX_j$ the $j$-th column of $\matX$ and by $[n]$ the set $\{1,2,\dots, n\}$. In addition, we denote by $\y$ the $m$-dimensional target vector. The mean value of $\y$, also referred to as \emph{mean target value}, is denoted by $\mu$. The tuple $(\matX, \y)$ contains all relevant information for a prediction task and we call such a tuple \emph{design matrix pair} or for simplicity, design matrix.
We focus on the typical binary classification task, i.e., assuming an input target vector $\y\in{\{0,1\}^m}$. The analysis can be extended directly to the regression task. Now we are ready to define target encodings.
\begin{definition}[Target Encodings]\label{def:target_enc}
Given $(\matX, \y)$ and an integer $j\in{[n]}$ so that the $j$-th column of $\matX$ is categorical, it follows that target encoding is a function $f_{(\matX_j, \y)}: \CC_j \to \RR$.
\end{definition}
From now on, we write $f$ instead of $f_{(\matX_j, \y)}$ for notation convenience. It is important to note that we allow $f$ to depend on the input dataset. Moreover, we say that $f$ is defined (or fitted) on $(\matX,\y)$ to explicitly specify the input data used on the definition of $f$.
A very common example of target encoding is the \emph{mean target encoding}. That is, assume that the $j$-th column of $\matX$ is a categorical containing values/levels in $\CC_j=\{L_1, L_2, \dots , L_k\}$. The mean target encoding $\mu_j$ of the $j$-th column is defined as follows: $\mu_j$ support on $\CC_j$ and for any $L\in \CC_j$,  
\begin{equation}\label{eq:mean_target_enc}
\mu_j(L)=\frac{1}{N}\sum_{i=1}^{m} y_i \mathbbm{1}_{\matX_{i,j} = L}
\end{equation}
where $N$ equals to the number of occurrences of $L$ in the $j$-th column of $\matX$ and $\mathbbm{1}_{\text{pred}}$ is the indicator function, i.e., equals to $1$ if pred is true, otherwise equals to zero. In words, mean target encodings are roughly defined as the mean target value of any level of the categorical (group).
In general, any property of the target values distribution of the group can be also extracted. For example, ML practitioners frequently use the minimum, maximum, standard deviation, kurtosis, percentiles of the target values in addition to the mean value. The main idea is to extract as much statistical information of the target distribution of the group as possible.
%
\subsubsection{Regularization of Target Encodings.}\label{sec:reg}
%
By definition, target encodings introduce target leakage and could lead to poor generalization performance, hence, target encoding regularization must always be used~\cite{leakage}.
In this section, we outline several regularization methods of target encodings. 
Extra caution on regularization should be given in the presence of high cardinality categoricals, i.e., categoricals with a large number of distinct levels as present in this competition. In fact, it is relatively easy to construct an example where the naive application of target encodings leads to severe overfitting. In order to exemplify this behavior, a minimal example is constructed by the authors of the `vtreat` package~\cite{vtreat}. CatBoost provides an implementation that handles these issues automatically, however, it is important for the modeler to better understand the general approaches that we outline next.
%
\paragraph{Smoothing / Empirical Bayes / Shrinkage of Mean Target Encoding.}\label{sec:reg:eb}
%
In the presence of high-cardinality categoricals, it is quite often the case that individual categorical levels appear only in a small number of samples. In such a scenario, the estimate of the mean target encoding doesn't generalize well due to the small number of samples used to calculate the statistics. Here, smoothing or shrinkage can be applied which have a similar effect as empirical Bayesian (EB) approaches~\cite{gelmanbda04}. Indeed, Empirical Bayesian conditional probabilities of a categorical can be understood as mean target encodings~\cite{encoding}.
In our notation, the EB regularized version of the mean target encoding is defined as
\begin{equation}\label{eq:target_enc:eb}
\mu^{\text{EB}}_j(L):= \lambda(N)  \mu_j(L) + (1- \lambda (N)) \mu
\end{equation}
where $N$ equals to the number of occurrences of $L$ in the $j$-th column of $\matX$ and $\lambda(n)$ is a monotonically increasing function on $n$ bounded between $0$ and $1$. A common choice of practitioners for $\lambda$ is  $\lambda (n) =\frac1{1+\exp(- ((n - l) / \sigma)}$ which is a $s$-shaped function with a value of $0.5$ for $n=l$ and $\sigma$ representing the steepness~\cite[Equation~$4$]{encoding}. Thus, Equation~\ref{eq:target_enc:eb} is a smoothed version of the mean target encoding.
%
\paragraph{Bootstrapping / rolling mean.}\label{sec:reg:bootstrap}
%
Bootstrapping is another approach to regularized target encodings. A specific instance of bootstrapping and target encodings is implemented in CatBoost~\cite{catboost}.
CatBoost uses a bootstrapping rolling mean approach to reduce overfitting while utilizing the whole training dataset for estimating the target encodings. In a nutshell, CatBoost performs a random permutation on the rows of $\matX$ and for the $i$-th row of $\matX$ (with respect to the random permutation) the mean target encoding is computed using only the rows up to the $(i-1)$-row. Namely, CatBoost averages several independent random permutations and, moreover, adds a shrinkage prior to the global mean.
To sum up, CatBoost performs categorical encoding for a level $L\in{\CC_j}$ as follows. For the $i$-th row and a fixed permutation of the rows, CatBoost computes
\[\mu_j^{\text{Cat}}(L,i) := \lambda \mu_j\left(L; (\matX_{1:(i-1),j}, \y_{1:(i-1)} \right) + (1-\lambda) \mu\]
where $\mu$ is the mean target value and $\lambda$ is a smoothing hyper-parameter.
%
\section{Winning Model and Additional Experiments}\label{sec:winning-model}
%
In this section, we describe the winning solution in more detail and present additional experimental results that better explain the critical choices made to achieve our result.
%
\subsection{Feature Engineering}\label{sec:feature-engineering}
We have extracted a range of features from the message log representing the training data. The following features focus on different aspects of the data (time/activity, message, game).

\begin{multicols}{2}
\begin{enumerate}
    \item \verb t_min: day of first message
    \item \verb t_max: day of last message
    \item \verb t_days: number of days with messages
    \item \verb|t_dur|: \verb|t_max| - \verb|t_min| 
    \item \verb t_active: \verb|t_days| / \verb|t_dur| 
    \item \verb m_total:  total number of characters for all messages
    \item \verb m_max: max number of character per message
    \item \verb m_med: medium number of character per message
    \item \verb g_n_mes: total number of messages
    \item \verb g_n: number of games
    \item \verb g_top: game with most messages
    \item \verb g_chat: number of messages for game just chatting
    \item \verb|g_top_freq|: fraction of messages for \verb|g_top| 
\end{enumerate}
\end{multicols}

All features have been computed as user- and channel-features. This can be done efficiently by aggregating the messages on either the user or channel id. Features such as the number of days between the first and last message (t\_dur) are computed for individual channel-user combinations.

In addition the following features have been added:

\begin{multicols}{2}
\begin{enumerate}
    \item \verb uid: user id 
    \item \verb n_channel: number of channels per user
    \item \verb u_group [low, normal, high]: user activity
    \item \verb cid: channel id 
    \item \verb n_user: number of user per channel
    \item \verb c_group [low, normal, high]: channel activity
\end{enumerate}
\end{multicols}

We show in Section \ref{sec:experiments} that only a small subset of all features is needed to achieve competitive performance.

%

%
%
\begin{figure}
 \centering
 \subfloat[Top features (train)]{
  \includegraphics[width=0.45 \textwidth]{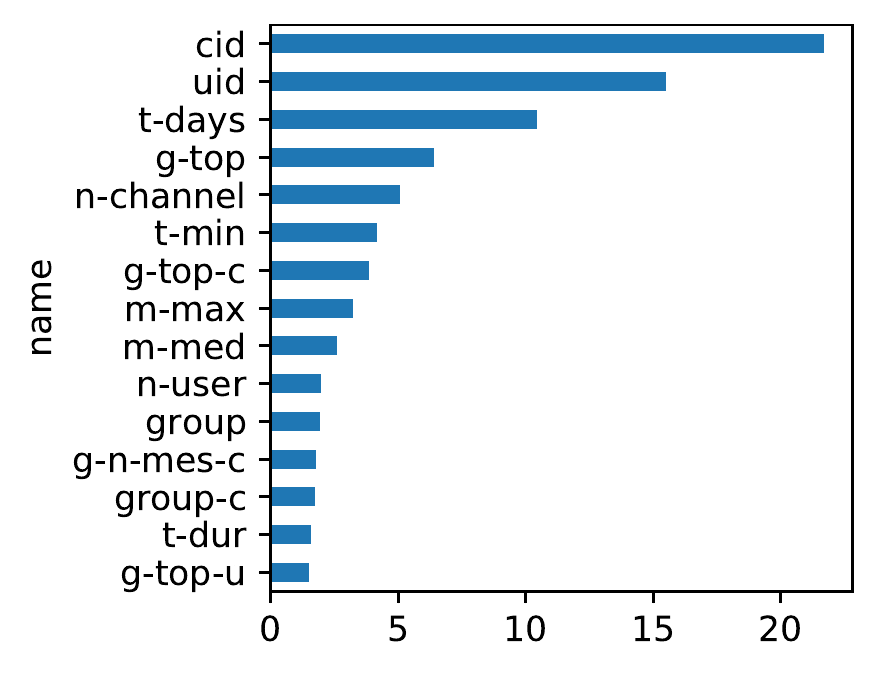}
}\qquad
 \subfloat[Top features (test)]{
  \includegraphics[width=0.45 \textwidth]{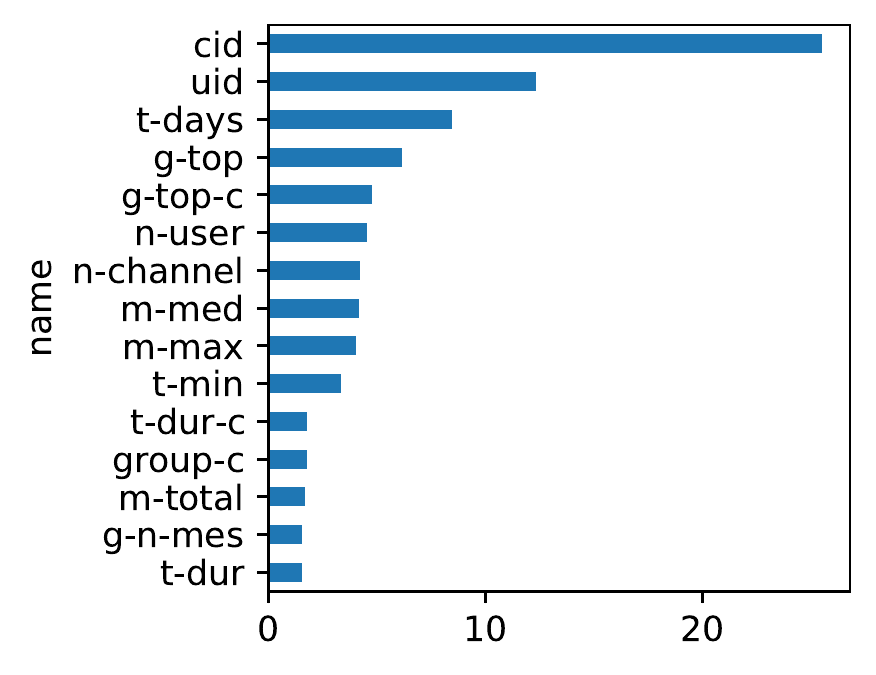}
}\\
\caption{CatBoost's feature importance show clear differences between train and our constructed test data
         that can be attributed to the group activity shift (especially pronounced for the uid feature). The top 10 test features have been selected for a simplified model~(table \ref{tab:model:stats}).}
\label{fig:importance}
\end{figure}
%
\subsection{Best Performing Model}
%
\paragraph{Model definition.} The model is based on the CatBoost library version $0.23.1$. The loss function is set to be \emph{logistic loss}, also known as cross-entropy loss. Training of the model is stopped early based on the performance on our custom validation set using the autostop capabilities of CatBoost (`od\_type` set to "Iter" and `od\_wait` set to $20$). The best model is selected automatically by setting use\_best\_model=True.
The top performing submission is a single CatBoost model trained with the following hyper-parameters: 'l2\_leaf\_reg': 64, 'learning\_rate': 0.08, 'threshold': 0.167, 'depth': 9, 'random\_strength': 0.5, 'max\_ctr\_complexity': 2. These parameters have been manually selected on our constructed test set.
\begin{table}[ht]
\scriptsize
\centering
\caption{Leaderboard results of the competition. Our submission  voyTECH ranked first with a clear lead to the second best approach.}\label{tab:leaderboard}
\begin{tabular}{|c|c|c|}
\toprule
\textbf{Rank} & \textbf{Teamname} & \textbf{Test F1 score} \\ \midrule
\textbf{1.}   & voyTECH           & 0.3433                 \\ \hline
\textbf{2.}   & CoolStoryBob      & 0.2647                 \\ \hline
\textbf{3.}   & ItsBoshyTime      & 0.2593                 \\ \hline
\textbf{4.}   & StinkyCheese      & 0.1422                 \\ \hline
\textbf{5.}   & Random Baseline   & 0.0741                 \\ \bottomrule
\end{tabular}
\end{table}
\paragraph{Cross validation Setup.}
Training and testing data come from a vastly different distribution, hence, a careful cross validation setup was crucial for model training and hyper-parameter tuning. It is given from the competition description that half of the users in the test set have no history and user-channel interactions are sampled uniformly from low, normal, and high activity levels. Therefore, our goal is to construct a validation set with similar properties. The validation set is constructed as follows: we sample in total 45,000 channel-user pairs (5,000 pairs per activity level pair group), ensuring that these pairs do not appear in the training set. These 45,000 channel-pairs are then duplicated in the validation set by modifying the user-id with an unknown identifier not present in the training set.
The training dataset is sub-sampled with a `max\_per\_group` parameter that restricts the number of samples per channel-user group.
%
\subsection{Additional Experiments}\label{sec:experiments}
%

\begin{figure}[ht]
 \centering
 \subfloat[Leaderboard Model]{
    \includegraphics[width=0.45 \textwidth]{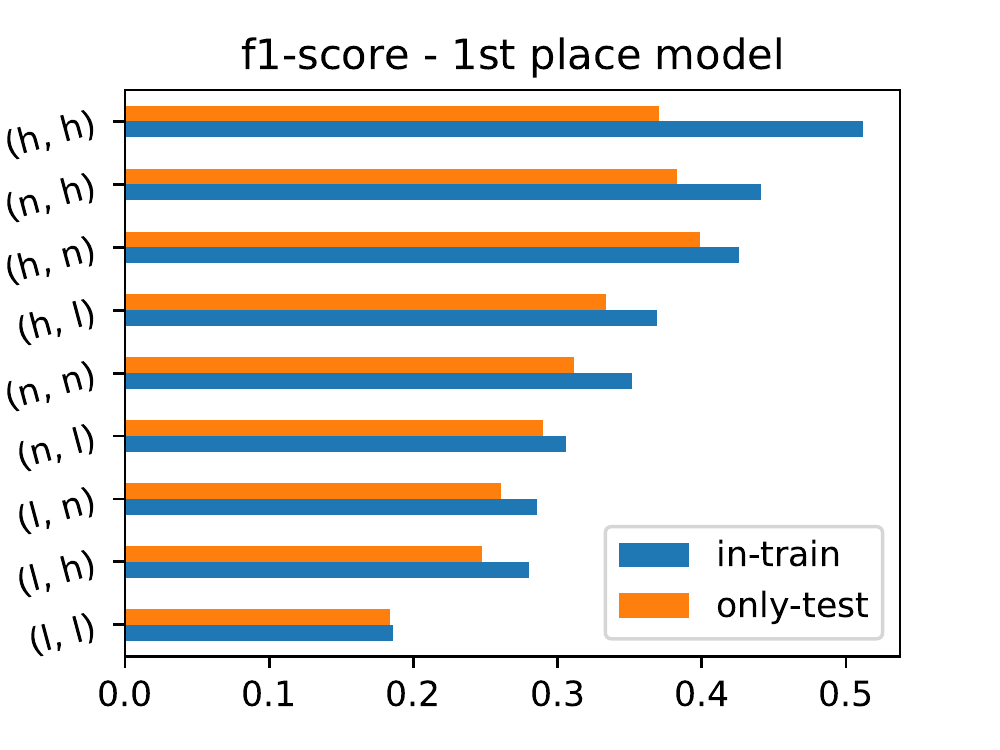}
  }\qquad
 \subfloat[Best Features Model]{
    \includegraphics[width=0.45 \textwidth]{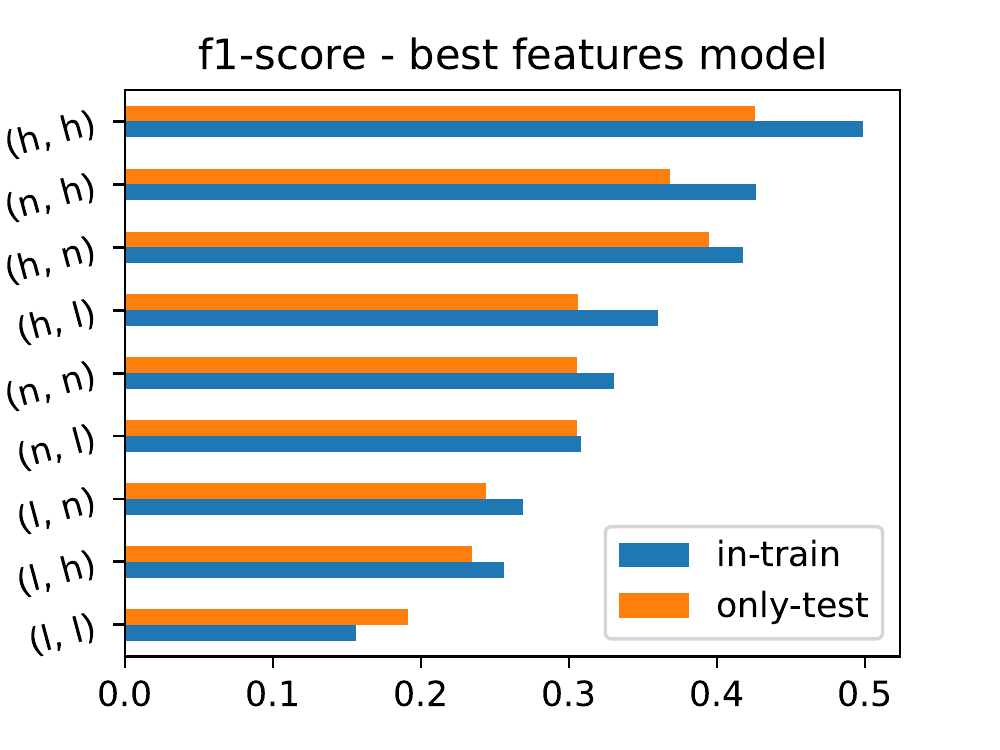}
}
\caption{Activity group specific F1-score. The differences between the full and reduced feature model is most pronounced for the highest (h,h) and lowest (l, l) activity groups.}
\label{group_performance}
\vspace{-4mm}
\end{figure}

We find that we can achieve strong performance~(Table \ref{tab:model:stats}) even with a very small subset of features ('uid', 'cid', 't\_days', 'g\_top', 'g\_top\_c', 'n\_user', 'n\_channel', 'm\_med', 'm\_max', 't\_min')
 selected by using CatBoot's feature importance~(Figure~\ref{fig:importance}). The sub-sampling of the training data decreases the model performance (Figure \ref{group_performance}) and the interaction between features (max\_ctr\_compl $\geq 2$) is important. However, the memory and run-time requirements increase dramatically when max\_ctr\_compl $> 2$. We therefore had to trade higher interactions against more training samples, which were more critical for performance.

\begin{table}[h]
\caption{Model performance with respect to train data size, features, and hyper-parameter (random-strength=0.5, threshold=0.167, l2-leaf-reg=64 and depth=9 are the same for all rows).}\label{tab:model:stats}
\centering
\scriptsize
\begin{tabular}{|l|c||c|c|c|c|c|}
    \toprule
f1 leaderboard & f1 mytest &train data & features     & max\_ctr\_compl &  lr   \\ \midrule
0.3433         & 0.3522    &16m        & all          & 2               &  0.15 \\ \hline
0.3382         & 0.3505    &8m         & all          & 2               &  0.08 \\ \hline
0.3345         & 0.3490    &16m        & top-features & 2               &  0.15 \\ \hline
0.3329         & 0.3417    &full       & top-features & 1               &  0.08 \\ \hline
0.3322         & 0.3421    &16m        & top-features & 1               &  0.15 \\ \bottomrule
\end{tabular}
\end{table}
%


\subsection{Conclusions \& Further Work}

In this work we showed that modeling user activity is more powerful than direct modeling of content when encoded properly and combined with a suitable optimization approach. We also introduced the connection between target encodings and boosting trees in the context of high cardinality categoricals and highlighted differences in the two popular boosting tree implementations CatBoost and LightGBM. 
We plan to conduct further experiments that compare Boosting Trees to Factorization Machines~\cite{rendle2010factorization},
a model that has been used successfully to model user activity in an earlier Discovery Challenge~\cite{bayer2013factor}.

\section*{Acknowledgement}

We would like to thank the organisers for the interesting competition and their support with the TIRA platform. From Palaimon we want to thank Phoebe Kuhn for her help with the manuscript and Alexander Pisarenko for his relentless work to provide us with a reliable and scalable infrastructure.



%
%
%
%

\medskip
{\scriptsize
\bibliographystyle{splncs04}
\bibliography{references}
}

\end{document}